\pdfoutput=1

\documentclass[11pt,a4paper]{article}

\usepackage{times}
\usepackage{latexsym}

\usepackage[acceptedWithA]{tacl2021v1}

\usepackage[T1]{fontenc}

\usepackage[utf8]{inputenc}

\usepackage{microtype}

\usepackage{inconsolata}

\usepackage{soul}

\usepackage{graphicx}
\usepackage{booktabs}
\usepackage{multirow}
\usepackage{multicol}
\usepackage{amssymb}
\usepackage[capitalise]{cleveref}
\usepackage{subcaption}

\Crefname{section}{Sec.}{Secs.}

\usepackage{pgfplots}
\usepackage[edges]{forest}
\definecolor{paired-light-blue}{RGB}{198, 219, 239}
\definecolor{paired-dark-blue}{RGB}{49, 130, 188}
\definecolor{paired-light-orange}{RGB}{251, 208, 162}
\definecolor{paired-dark-orange}{RGB}{230, 85, 12}
\definecolor{paired-light-green}{RGB}{199, 233, 193}
\definecolor{paired-dark-green}{RGB}{49, 163, 83}
\definecolor{paired-light-purple}{RGB}{218, 218, 235}
\definecolor{paired-dark-purple}{RGB}{117, 107, 176}
\definecolor{paired-light-gray}{RGB}{217, 217, 217}
\definecolor{paired-dark-gray}{RGB}{99, 99, 99}
\definecolor{paired-light-pink}{RGB}{222, 158, 214}
\definecolor{paired-dark-pink}{RGB}{123, 65, 115}
\definecolor{paired-light-red}{RGB}{231, 150, 156}
\definecolor{paired-dark-red}{RGB}{131, 60, 56}
\definecolor{paired-light-yellow}{RGB}{231, 204, 149}
\definecolor{paired-dark-yellow}{RGB}{141, 109, 49}
\definecolor{light-green}{RGB}{118, 207, 180}
\definecolor{raspberry}{RGB}{228, 24, 99}

\usepackage[misc]{ifsym}

\newcommand{\camready}[1]{}

\newcommand{\com}[1]{}

\title{Efficient Methods for Natural Language Processing: A Survey}

\author{
    Marcos Treviso$^{1}$\thanks{~~Equal contribution. \Letter\  \texttt{marcos.treviso@tecnico.pt}},
    Ji-Ung Lee$^{2*}$,
    Tianchu Ji$^{3*}$, 
    Betty van Aken$^{4}$, 
    Qingqing Cao$^{5}$,
    \\
    \bf 
    Manuel R. Ciosici$^{6}$,
    Michael Hassid$^{7}$,
    Kenneth Heafield$^{8}$,
    Sara Hooker$^{9}$,
    \\
    \bf
    Colin Raffel$^{10}$,
    Pedro H. Martins$^{1,11}$,
    André F. T. Martins$^{1,11}$, 
    Jessica Zosa Forde$^{12}$,
    \\
    \bf 
    Peter Milder$^{3}$,
    Edwin Simpson$^{13}$,
    Noam Slonim$^{14}$,
    Jesse Dodge$^{15}$,
    Emma Strubell$^{15,16}$,
    \\
    \bf
    Niranjan Balasubramanian$^{3}$, 
    Leon Derczynski$^{5,17}$,
    Iryna Gurevych$^{2}$, 
    Roy Schwartz$^{7}$
    \\
    $^{1}$IST/U. of Lisbon \& Instituto de Telecomunicações,
    $^{2}$Technical University of Darmstadt, \\
    $^{3}$Stony Brook University,
    $^{4}$Berliner Hochschule für Technik,  
    $^{5}$University of Washington, \\
    $^{6}$University of Southern California, 
    $^{7}$The Hebrew University of Jerusalem, \\
    $^{8}$University of Edinburgh,
    $^{9}$Cohere For AI, 
    $^{10}$University of North Carolina at Chapel Hill, \\
    $^{11}$Unbabel, 
    $^{12}$Brown University, 
    $^{13}$University of Bristol, 
    $^{14}$IBM Research,  \\
    $^{15}$Allen Institute for AI,
    $^{16}$Carnegie Mellon University,
    $^{17}$IT University of Copenhagen
}

\begin{document}
\maketitle
\begin{abstract}
Recent work in natural language processing (NLP) has yielded appealing results from scaling model parameters and training data; however, using only scale to improve performance means that resource consumption also grows. Such resources include data, time, storage, or energy, all of which are naturally limited and unevenly distributed. This motivates research into \textit{efficient} methods that require fewer resources to achieve similar results. This survey synthesizes and relates current methods and findings in efficient NLP. We aim to provide both guidance for conducting NLP under limited resources, and point towards promising research directions for developing more efficient methods.
\end{abstract}

\section{Introduction}
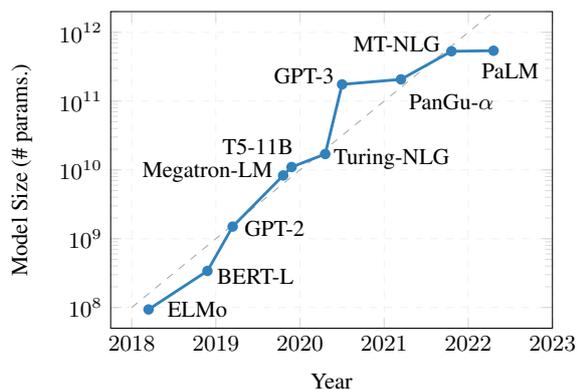
\begin{figure}[t]
    \centering
    \footnotesize
    \resizebox{\columnwidth}{!}{%
    \begin{tikzpicture}
    \begin{axis}[
        width=\columnwidth,
        height=6cm,
        xlabel={Year},
        ylabel={Model Size (\# params.)},
        grid=major,
        major grid style={line width=.1pt, draw=gray!10},
        xmin=2018-0.25, 
        xmax=2023,
        ymin=5e7, 
        ymax=2e12,
        xtick={2018,2019,2020,2021,2022,2023},
        x tick label style={
            /pgf/number format/.cd,
                fixed,
                set thousands separator={},
            /tikz/.cd
        },
        ytick={1e8, 1e9, 1e10, 1e11, 1e12, 1e13},
        y tick label style={
            /pgf/number format/.cd,
                fixed,
                sci,
            /tikz/.cd
        },
        every x tick/.style={black!15},
        every y tick/.style={black!15},
        ymode=log,
        grid style=dashed,
    ]
    
    \addplot[
    dashed,
    color=gray!70,
    ]
    coordinates {
    (2018, 1e8) 
    (2019, 1e9) 
    (2020, 1e10) 
    (2021, 1e11) 
    (2022, 1e12)  
    (2023, 1e13)
    };
    
    \addplot[
    solid,
    color=paired-dark-blue,
    line width=0.4mm,
    mark=*,
    mark size=1.5pt,
    ]
    coordinates {
    (2018.2, 93.6e6) 
    (2018.9, 340e6) 
    (2019.2, 1.5e9) 
    (2019.8, 8.3e9) 
    (2019.9, 11e9) 
    (2020.3, 17e9) 
    (2020.5, 175e9) 
    (2021.2, 207e9) 
    (2021.8, 530e9) 
    (2022.3, 540e9)
    };

    \draw[] (axis cs:2018.2, 1e8) + (0.58,-0.05) node {ELMo};
    \draw[] (axis cs:2018.9, 6e8) + (0.57,-9pt) node {BERT-L};
    \draw[] (axis cs:2019.2, 2e9) + (0.5,-4pt) node {GPT-2};
    \draw[] (axis cs:2019.8, 1e10) + (-0.9,-1pt) node {Megatron-LM};
    \draw[] (axis cs:2019.9, 1.3e10) + (-0.4,6pt) node {T5-11B};
    \draw[] (axis cs:2020.3, 2e10) + (0.79,-4pt) node {Turing-NLG};
    \draw[] (axis cs:2020.5, 1.2e11) + (-0.45,8pt) node {GPT-3};
    \draw[] (axis cs:2021.2, 1.3e11) + (0.6,-3.5pt) node {PanGu-$\alpha$};
    \draw[] (axis cs:2021.8, 530e9) + (-0.65,3pt) node {MT-NLG};
    \draw[] (axis cs:2022.3, 540e9) + (0.2,-8pt) node {PaLM};
    
    \end{axis}
    \end{tikzpicture}
    }
    \vspace{-0.7cm}
    \caption{Exponential growth in the number of parameters in pretrained language models. Adapted from \citet{Lakim2022AHA}.} \label{fig:large_pretrained_models}
    \vspace{-0.25cm}
\end{figure}

Scaling has become a key ingredient in achieving %
state-of-the-art performance in NLP (\cref{fig:large_pretrained_models}), %
as recent research suggests that some capabilities only emerge once models grow beyond a certain size~\citep{wei_emergent_2022}.
However, despite the merits of scaling, it poses key challenges to making these breakthroughs accessible in resource constrained environments \citep{ahmed2020dedemocratization}; %
 in having a non-negligible environmental impact~\citep{Strubell:2019,Schwartz:2020,derczynski2020power, patterson2021carbon, wu2022sustainable}; and in complying with hardware constraints~\citep{thompson2020computational}.
To tackle these limitations, there has been renewed focus around research that seeks to improve model \textit{efficiency}.

\paragraph{Definition} 
Efficiency is characterized by the relationship between resources going into a system and its output, with a more efficient system producing the same output with fewer resources. 
\citet{Schwartz:2020} formalize efficiency as the cost of a model in relation to the results it produces: $\mathrm{Cost}(R) \propto E \cdot D\cdot H,$ %
i.e., the $\mathrm{Cost}(\cdot)$ of producing a certain NLP ($R$)esult as proportional to three (non-exhaustive) factors: (1) The cost of model execution on a single ($E$)xample, (2) the size of the ($D$)ataset %
and (3) the number of training runs required for ($H$)yperparameter tuning.
Here we take a different approach, and consider the role efficiency plays across the different steps in the NLP pipeline, by providing a detailed overview of efficiency methods specific to NLP
 (\cref{fig:overview}). 

\paragraph{Scope of this survey} 

We address this work to two groups of readers: 
(1) Researchers from all fields of NLP working with limited resources; and 
(2) Researchers interested in improving the state of the art of efficient methods in NLP. 
Each section concludes with a discussion of limitations, open challenges, and possible future directions of the presented methods. We start by discussing methods to increase \textit{data} efficiency (\cref{sec:data}), and continue with methods related to \textit{model design} (\cref{sec:model_design}). We then consider efficient methods for the two typical training setups in modern NLP: \textit{pre-training }(\cref{sec:pretraining}) and \textit{fine-tuning} (\cref{sec:finetuning}). We then discuss methods for making \textit{inference} more efficient (\cref{sec:inference}). 
While we mainly focus on algorithmic approaches, we provide appropriate pointers regarding \textit{hardware} that are connected to the scale at which we expect to deploy a model (\cref{sec:hardware}). We then discuss how to quantify efficiency and what factors to consider during \textit{evaluation} (\cref{sec:evaluation}), and finally---how to efficiently decide upon the \textit{best suited model} (\cref{sec:model_selection}). %

\begin{figure*}[!htb]
    \scriptsize
    \centering
    \resizebox{0.9\textwidth}{!}{%
    \begin{minipage}[b]{0.5\linewidth}
    \centering
    \begin{forest}
        for tree={
            forked edges,
            grow'=0,
            draw,
            rounded corners,
            node options={align=center,},
            text width=2cm,
            s sep=4pt,
            calign=child edge, 
            calign child=(n_children()+1)/2,
            l sep=7.5pt,
        },
        [, phantom
            [Data \S\ref{sec:data}, for tree={ data}
                [Filtering,  data
                    [\citet{mishra-sachdeva-2020-need,zhang2022opt}, data_work]
                ]
                [Curriculum Learning ,  data
                    [\citet{wan-etal-2020-self,press-etal-2021-shortformer,zhu-etal-2021-combining-curriculum-learning}, data_work]
                ]
                [Active Learning ,  data
                    [\citet{ein-dor-etal-2020-active,yuan-etal-2022-adapting,lee-klie-gurevych-2022}, data_work]
                ]
            ]
            [Model Design \S\ref{sec:model_design}, for tree={fill=red!45,model}
                [Compres. Attention, model
                    [
                    Transformer-XL~\citep{dai2019transformer}; 
                    $\infty$-former~\citep{martins2022infty}, model_work]
                ]
                [Fast Attention, model
                    [
                    Reformer~\citep{kitaev2020reformer}; \\ Performer~\citep{choromanski2020rethinking}, model_work]
                ]
                [Sparse \\Modeling, model
                    [Switch Transf.~\citep{fedus2021switch}; Sparsefinder~\citep{treviso-etal-2022-predicting}, model_work]
                ]
                [Parameter Efficiency, model
                    [
                    ALBERT~\citep{lan2019albert};
                    Perceiver~\citep{jaegle2021perceiver};
                    , model_work]        
                ]
                [Retrieval-based, model
                    [$k$NN-LM~\citep{khandelwal2019generalization}; RETRO~\citep{borgeaud2021improving}, model_work]  
                ]
            ]
            [Pre-training  \S\ref{sec:pretraining}, pretraining
                [Decoder only, pretraining
                    [GPT-3~\citep{brown2020language}; PaLM~\citep{palm}, pretraining_work]
                ]
                [Encoder only, pretraining
                    [BERT~\citep{Devlin:2019}; ELECTRA~\citep{Clark:2020}, pretraining_work]
                ]
                [Encoder-Decoder, pretraining
                    [T5~\citep{raffel2019exploring}; BART~\citep{lewis-etal-2020-bart}, pretraining_work]
                ]
            ]
        ]
    \end{forest}
    \end{minipage}%
    \hspace*{0.4cm}
    \begin{minipage}[b]{0.5\linewidth}
    \begin{forest}
        for tree={
            forked edges,
            grow'=0,
            draw,
            rounded corners,
            node options={align=center,},
            text width=2cm,
            s sep=4pt,
            calign=edge midpoint,
            l sep=7.5pt,
        },
        [, phantom
            [Fine-tuning \S\ref{sec:finetuning}, finetuning
                [Parameter-Efficiency, finetuning
                    [Adapters~\citep{houlsby2019parameter}; \\ LoRA~\citep{hu2022lora}, finetuning_work]
                ]
                [Multi-task Learning, finetuning
                    [T5~\citep{raffel2019exploring}; \\ (IA)$^3$~\citep{liu2022few}, finetuning_work]
                ]
                [Zero-shot Learning, finetuning
                    [T0~\citep{sanh2022multitask}; \\ FLAN~\citep{wei2022finetuned}, finetuning_work]
                ]
                [Prompting,  finetuning
                    [GPT-3~\citep{brown2020language}; PET~\citep{schick-schutze-2021-just}, finetuning_work]
                ]
            ]
            [Inference \& Compression \S\ref{sec:inference}, for tree={inference}
                [Pruning, inference
                    [Magnitude P.~\citep{gordon_compressing_2020}; Movement P. ~\citep{sanh_movement_2020}, inference_work]
                ]
                [Distillation, inference
                    [TinyBERT~\citep{jiao-etal-2020-tinybert}; MobileBERT~\citep{sun-etal-2020-mobilebert}, inference_work]
                ]
                [Adaptive Computation, inference
                    [Tied Transf.~\citep{Dabre:2020}; \\ 
                    Depth-Adaptive Transf. \citep{Elbayad:2020}, inference_work]
                ]
                [Quantiza\-tion, inference
                    [8-bit Transf.~\citep{bhandare_efficient_2019};\\
                    Q-BERT~\citep{shen_q-bert_2020}, inference_work]
                ]
            ]
            [Hardware Utilization \S\ref{sec:hardware}, for tree={hardware}
                [Libraries, hardware
                    [DeepSpeed~\citep{ren_zero-offload_2021}; \\ bits\&bytes~\citep{dettmers2022bit}, hardware_work]
                ]
                [Specialized Hardware, hardware
                    [\citet{li_efficient_2021,qu_dota_2022,tambe_edgebert_2021}, hardware_work]
                ]
                [Edge Devices, hardware
                    [SqueezeBERT~\citep{iandolaSqueezeBERTWhatCan2020a};\\ ProFormer~\citep{sankarProFormerOnDeviceLSH2021}, hardware_work]
                ]
            ]
        ]
    \end{forest}
    \end{minipage}
    }
    \caption{Typology of efficient NLP methods. 
    }
    \label{fig:typo-efficient-nlp}
    \vspace{-0.25cm}
\end{figure*}
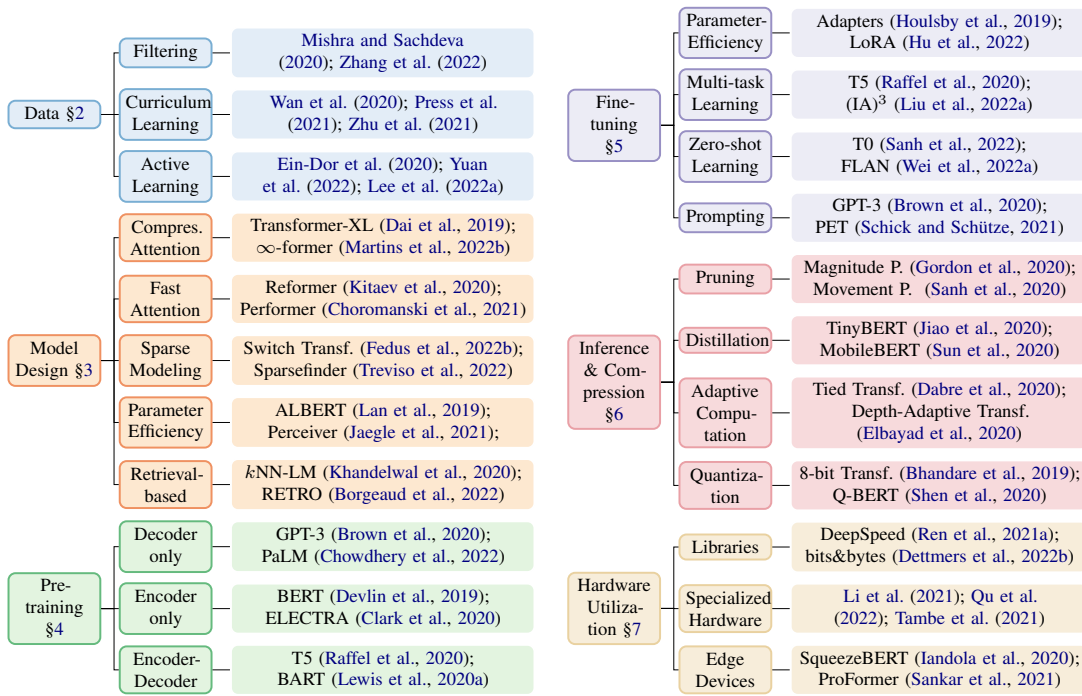

To guide the reader, \cref{fig:typo-efficient-nlp} presents a typology of efficient NLP methods considered in this survey.

\com{Here, we adopt the equation to attribute methods to their respective stage in the NLP model-building pipeline (\cref{fig:overview}): 
\begin{equation}
    \mathrm{Cost}(\mathcal{R}) \propto I\cdot [ U\cdot (D_{P}\cdot H_{P}+D_{F}\cdot H_{F}) + c ], \label{eq:cost}
\end{equation}
where we distinguish between model ($I$)nference and model ($U$)pdate.
We further distinguish between ($P$)re-training a model on large amounts of text in a self-supervised fashion, ($F$)ine-tuning it to a specific downstream task, and the ($c$)ost of deploying the model.
Since using a single cost indicator can be misleading~\citep{dehghani2021}, we further extend $R$ to a set of possible ($\mathcal{R}$)esults that may need to be traded-off.%

}

\com{For readers interested in a specific aspect of efficiency, the following list connects \cref{eq:cost} to their respective sections:

\noindent
($I$)nference \S\ref{sec:model_design}, \S\ref{sec:inference}, \S\ref{sec:hardware}; ($U$)pdates \S\ref{sec:curriculum}, \S\ref{sec:model_design}, \S\ref{sec:finetuning}, \S\ref{sec:pruning}, \S\ref{sec:quantization}, \S\ref{sec:hardware};($D$)ata \S\ref{sec:data}; ($H$)yperparameter Tuning \S\ref{sec:model_selection}; Deployment ($c$)ost  \S\ref{sec:pruning}, \S\ref{sec:hardware}; ($\mathcal{R}$)esults \S\ref{sec:evaluation}.}

\section{Data} \label{sec:data}
Data efficiency is improved by using fewer training instances, or by making better use of available instances. Fixed compute budgets motivate balancing model size and training data size, especially during pre-training~\citep{Hoffmann2022}.%

\subsection{Filtering}
Improving \textit{data quality} can boost performance while reducing training costs during pre-training and fine-tuning.
For instance, \citet{lee-etal-2022-deduplicating} showed that removing duplicates in pre-training increases training efficiency, giving equal or even better model performance compared to using all data. %
\citet{zhang2022opt} used MinhashLSH~\citep{leskovec2020mining} to remove duplicates while developing OPT.
De-duplication can lead to substantially reduced computation cost, especially in cases with abundant pre-training data but limited compute budget~\citep{Hoffmann2022}.

Similar observations have been made for fine-tuning. 
For instance, \citet{mishra-sachdeva-2020-need} found---via adversarial filtering~\cite{Zellers:2018}---a subset of only $\sim$2\% of the SNLI data~\citep{bowman-etal-2015-large} that leads to performance comparable to using the full corpus. %
While such filtering approaches are useful for mitigating biases~\citep{LeBras:2020}, they may not always serve as tools to filter existing datasets, as these often suffer from insufficient training data. %

\tikzset{%
    modelnode/.style = {
        align=center,
        rounded corners=3pt,
        inner sep=0.21cm,
        fill=paired-light-orange!50,
        text width=2cm,
        draw=black,
        line width=0.2mm
    },
    pretrainingnode/.style = {
        align=center,
        rounded corners=3pt,
        inner sep=0.21cm,
        fill=paired-light-green!50,
        text width=2cm,
        draw=black,
        line width=0.2mm
    },
    finetuningnode/.style = {
        align=center,
        rounded corners=3pt,
        inner sep=0.21cm,
        fill=paired-light-purple!50,
        text width=2cm,
        draw=black,
        line width=0.2mm
    },
    inferencenode/.style = {
        align=center,
        rounded corners=3pt,
        inner sep=0.21cm,
        fill=paired-light-red!35,
        text width=2cm,
        draw=black,
        line width=0.2mm
    },
    datanode/.style = {
        align=center,
        rounded corners=3pt,
        inner sep=0.25cm,
        fill=paired-light-blue!50,
        draw=blue!0,
        text width=2.4cm,
        draw=black,
        line width=0.3mm,
    },
    selectionnode/.style = {
        align=center,
        rounded corners=3pt,
        inner sep=0.21cm,
        fill=paired-light-gray!10,
        text width=2cm,
        draw=black,
        line width=0.2mm
    },
}

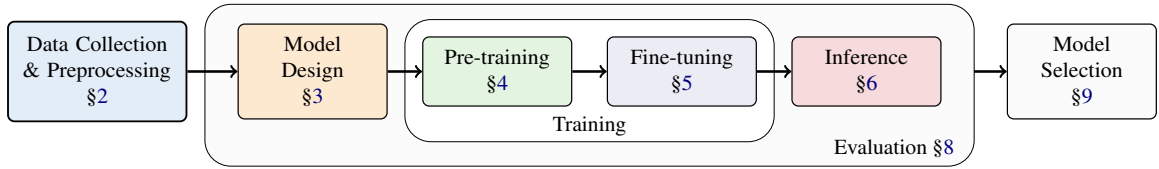
\begin{figure*}
    \small
    \centering
    \resizebox{0.95\textwidth}{!}{%
    \begin{tikzpicture}[auto, node distance=2cm]
        \draw[color=black, fill=paired-light-gray!10, rounded corners=12pt, very thin] (1.75, -1.55) rectangle (14.25, 1.1);
        \draw[color=black, fill=white, rounded corners=12pt, very thin] (5, -1.15) rectangle (11, 0.85);
        
        \node[datanode] (data) at (0, 0) {
            Data Collection \& Preprocessing \\ \S\ref{sec:data}
        };
        
        \node[modelnode] (modeldesign) at (3.5, 0) {
            Model Design \\ \S\ref{sec:model_design}
        };
        \node[pretrainingnode] (pretraining) at (6.5, 0) {
            Pre-training \\ \S\ref{sec:pretraining}
        };
        \node[finetuningnode] (finetuning) at (9.5, 0) {
            Fine-tuning \\ \S\ref{sec:finetuning}
        };
        \node[inferencenode] (inference) at (12.5, 0) {
            Inference \\ \S\ref{sec:inference}
        };
        \node[selectionnode] (modelsel) at (16, 0) {
            Model Selection \\ \S\ref{sec:model_selection}
        };
        
        \draw[->,very thick] (data) to (modeldesign);
        \draw[->,very thick] (modeldesign) to (pretraining);
        \draw[->,very thick] (pretraining) to (finetuning);
        \draw[->,very thick] (finetuning) to (inference);
        \draw[->,very thick] (14.25, 0) -- (modelsel);
        
        \node (desc1) at (8, -0.9) {Training};
        \node (desc2) at (12.95, -1.25) {Evaluation \S\ref{sec:evaluation}};

    \end{tikzpicture}
    }
    \vspace*{-0.2cm}
    \caption{Schematic overview of the efficient NLP stages covered in this paper, starting with data collection and model design, followed by training and inference, and ending with evaluation and model selection. 
    Notably, the training stage is divided into two parts: pre-training, which aims to learn generalizable parameters, and fine-tuning, which optimizes these parameters for specific downstream tasks. 
    }
    \label{fig:overview}
\end{figure*}

\tikzset{%
    parent/.style =          {align=center,text width=2cm,rounded corners=3pt, line width=0.3mm, fill=gray!10,draw=gray!80},
    child/.style =           {align=center,text width=2.3cm,rounded corners=3pt, fill=blue!10,draw=blue!80,line width=0.3mm},
    grandchild/.style =      {align=center,text width=2cm,rounded corners=3pt},
    greatgrandchild/.style = {align=center,text width=1.5cm,rounded corners=3pt},
    greatgrandchild2/.style = {align=center,text width=1.5cm,rounded corners=3pt},    
    referenceblock/.style =  {align=center,text width=1.5cm,rounded corners=2pt},
    data/.style =           {align=center,text width=1.2cm,rounded corners=3pt, fill=paired-light-blue!50,draw=paired-dark-blue!65,line width=0.3mm},   
    data_work/.style =      {align=center, text width=4.4cm,rounded corners=3pt, fill=paired-light-blue!50,draw=blue!0,line width=0.3mm},  
    model/.style =           {align=center,text width=1.2cm,rounded corners=3pt, fill=paired-light-orange!50,draw=paired-dark-orange!65,line width=0.3mm},   
    model_work/.style =      {align=center,text width=4.4cm,rounded corners=3pt, fill=paired-light-orange!50,draw=red!0,line width=0.3mm},    
    pretraining/.style =        {align=center,text width=1.2cm,rounded corners=3pt, fill= paired-light-green!50,draw=paired-dark-green!75,line width=0.3mm},   
    pretraining_work/.style =   {align=center,text width=4.4cm,rounded corners=3pt, fill= paired-light-green!50,draw= cyan!0,line width=0.3mm},      
    finetuning/.style =           {align=center,text width=1.2cm,rounded corners=3pt, fill= paired-light-purple!50,draw=paired-dark-purple!75,line width=0.3mm},   
    finetuning_work/.style =      {align=center,text width=4.3cm,rounded corners=3pt, fill= paired-light-purple!50,draw= orange!0,line width=0.3mm},        
    inference/.style =           {align=center,text width=1.2cm,rounded corners=3pt, fill= paired-light-red!35,draw=paired-light-red!90,line width=0.3mm},   
    inference_work/.style =      {align=center,text width=4.3cm,rounded corners=3pt, fill= paired-light-red!35,draw= magenta!0,line width=0.3mm},
    hardware/.style =           {align=center,text width=1.2cm,rounded corners=3pt, fill= paired-light-yellow!35,draw=paired-light-yellow!90,line width=0.3mm},   
    hardware_work/.style =      {align=center,text width=4.3cm,rounded corners=3pt, fill= paired-light-yellow!35,draw= magenta!0,line width=0.3mm},
}

\subsection{Active Learning}
Active learning aims to reduce the number of training instances.
In contrast to filtering, %
it is applied during data collection (instead of after) to only annotate the most helpful or useful instances for training~\citep{Settles2012,10.1145/3472291}.
To assess usefulness of an instance without knowing its actual label, one can use the model \textit{uncertainty}---assuming that labeling instances with the highest uncertainty is most helpful~\citep{lewis1994sequential,tang-etal-2002-active,gal2017deep,yuan-etal-2020-cold}; instance \textit{representativeness}---to maximize diversity of sampled instances while avoiding outliers~\citep{bodo2011active,sener2018active,gissin2019discriminative}; or a combination of both criteria \citep{kirsch2019batchbald, Ash2020Deep, margatina-etal-2021-active,Shoaib2022,Agarwal_2022_CVPR}.
Active learning has been successfully applied in machine translation (MT, \citealt{liu-etal-2018-learning}), language learning~\citep{lee-etal-2020-empowering}, entity linking~\citep{klie-etal-2020-zero}, and coreference resolution~\citep{li-etal-2020-active-learning,yuan-etal-2022-adapting}.
Despite its advantages, some open questions make active learning difficult to apply in practice. 
It remains unclear how model-based sampling impacts the performance of models using architectures different from that in sampling~\citep{lowell-etal-2019-practical, ein-dor-etal-2020-active}.
Also, selecting ``difficult'' instances may increase annotation cost and difficulty~\citep{settles2008active,lee-klie-gurevych-2022}.
Finally, it is prone to selection biases and can favor outliers~\citep{cortes2008sampling-bias, karamcheti-etal-2021-mind}.

\subsection{Curriculum Learning}\label{sec:curriculum}
Curriculum learning aims to find a data ordering that reduces the number of training steps required to achieve a target performance~\citep{elman1993learning,Bengio2009}. %
This method does not reduce dataset size, but does improve its utilization. Hence, it is a common approach for improving training efficiency in both pre-training and fine-tuning.
Many curriculum learning methods order instances by difficulty,  using heuristics such as sentence length. %
This has yielded improvements for transformer pre-training~\citep{press-etal-2021-shortformer,agrawal-etal-2021-role} as well as fine-tuning on tasks such as question answering~\cite{tay-etal-2019-simple}, MT~\citep{zhang-etal-2019-curriculum}, and others~\cite{xu-etal-2020-curriculum}.

A major challenge in curriculum learning is determining \textit{pace}, i.e., when to progress to more difficult instances. If not chosen carefully, curriculum learning can waste compute on ``easy'' instances.
To tackle this, work has investigated adaptive ordering strategies based on current model state, called \textit{self-paced learning}~\citep{kumar2010self}.
This has been successfully applied to improve performance in MT using model and data uncertainty~\citep{wan-etal-2020-self,zhou-etal-2020-uncertainty, zhao2020reinforced}, and in dialog generation with knowledge distillation~\citep{zhu-etal-2021-combining-curriculum-learning}. 
However, self-paced learning involves large training costs, and disentangling instance ordering from factors such as optimizer choice and batch size is non-trivial~\citep{Dodge:2020}.

\subsection{Estimating Data Quality}

In an era of ever larger datasets, auditing and estimating the quality of data is increasingly challenging. Datasets frequently present high levels of noise and misaligned instances ~\citep{kreutzer-etal-2022-quality}. Estimating data quality encompasses research efforts which propose better uncertainty estimates~\citep{baldock2021, souza2021, ethayarajh2022understanding} as well as analytical tools such as dataset cartography~\citep{swayamdipta-etal-2020-dataset}. Qualitative tools include documentation for datasets and model attributes ~\citep{gebru2021datasheets}.

\section{Model Design}  \label{sec:model_design}
Efficient model design covers architectural changes and adding new modules to accelerate training.%

\subsection{Improving Attention in Transformers}\label{sec:attention}
The transformer's self-attention mechanism has a quadratic dependency on sequence length which 
is not fully utilized by existing models \cite{Hassid:2022}. To reduce computational costs, efficient attention mechanisms for long sequences have been proposed~\citep{tay2020efficient}. 
Existing strategies include better using already processed segments via recurrence to connect multiple segments~\citep{dai2019transformer}, learning a network to compress a longer-term memory~ \citep{rae2019compressive}, separately modeling global and local attention~\citep{ravula_etc_2020},
and modeling long inputs as a continuous-time signal \citep{martins2022infty}.
Another line of research uses fixed attention patterns, where tokens attend to their immediate context (local attention) and possibly to a few global positions (global attention; \citealp{beltagy_longformer_2020,zaheer2020big, child_generating_2019}). %
Compared to using the full self-attention matrix, such approaches can scale linearly with the input length. 

Some methods learn attention sparsity patterns directly from data, e.g.~by grouping tokens into buckets, leading to a more accurate yet more expensive approximation of the full attention matrix \citep{kitaev2020reformer,DarasSMYRF2020,roy2021efficient}.
Instead of seeking better attention patterns, some strategies modify the attention \textit{mechanism} and derive low-rank approximations to the query-key matrices via reverse application of the kernel trick, resulting in linear time attention \citep{katharopoulos2020transformers,choromanski2020rethinking,RFA,zhai2021attention}.  
Recently, IO-aware attention mechanisms have been proposed, decreasing reads and writes to the attention matrix to GPU high-bandwidth memory~\citep{Dao2022-yl}.%

Despite various improvements in attention mechanisms, most of them struggle with very long sequences~\citep{tay2021long}.
S4~\citep{gu2021efficiently}, and its successors \citep{gupta2022diagonal,mehta2022long,gu2022parameterization}, suggest an alternative to transformers that alleviates the short memory problem and the quadratic bottleneck cost of self-attention by discretizing state space representations through parameterization of the state matrix.
More recently, Mega~\citep{ma2022mega} replaced the multi-headed transformer attention mechanism with a single-headed mechanism that receives contextualized vectors from a multidimensional exponential moving average module, and then splits the input into multiple fixed-length chunks to reduce the computation cost. 
Both S4 and Mega strongly outperform attention-based methods on all tasks of the Long Range Arena benchmark~\citep{tay2021long}, while increasing training speed by approximately 5x and reducing memory cost by about 15\% when compared to a standard transformer.
This success is attributed to their convolutional structure, which emphasizes nearby tokens and has a parameter count that grows sub-linearly with sequence length~\citep{li2022makes}.
\looseness=-1%

\subsection{Sparse Modeling}  

To leverage sparsity for efficiency, many models follow the mixture-of-experts (MoE) concept~ \citep{jacobs1991adaptive,shazeer2017outrageously,fedus2022review}, which routes computation through small subnetworks instead of passing the input through the entire model.
Relevant works on this line include GShard~\citep{Lepikhin2021GShardSG}, Switch Transformer~\citep{fedus2021switch}, and ST-MoE~\citep{zoph2022designing}, which replace the feed forward layers in transformers with MoE layers.
More recently, \citet{Rajbhandari2022} scaled transformers up by compressing and optimizing the usage of MoE.
Overall, MoE models have been shown to achieve strong performance across several NLP tasks while reducing the overall resource consumption (\cref{sec:evaluation}). %
For instance, GLaM~\citep{pmlr-v162-du22c} used only $\sim$$\frac{1}{3}$ of GPT-3's energy consumption (with additional hardware-based optimization), while \citet{Rajbhandari2022} reached a 5x reduction in terms of training cost.
However, MoE models have also exhibited training instabilities in practice, %
and may require architecture-specific implementation~\citep{zoph2022designing,mustafa2022multimodal}.

Another promising direction for exploiting sparse modeling is Sparsefinder~\citep{treviso-etal-2022-predicting}, which extends the Adaptively Sparse Transformer \citep{correia_adaptively_2019} to allow a more efficient attention mechanism by identifying beforehand the sparsity pattern returned by entmax attention---a sparse alternative to (dense) softmax attention~\citep{peters-etal-2019-sparse}.
Finally, sparsity can also be induced via modularity, e.g., by encapsulating task-specific parameters~\citep{ponti2022combining}.%

\subsection{Parameter Efficiency} 
Methods that reduce parameter count can reduce computational costs and memory usage.
One such approach is to share weights across layers of a model while maintaining the downstream task performance \citep{dehghani2018universal,lan2019albert}.
Besides sharing weights, Perceiver~\citep{jaegle2021perceiver} also minimizes the computational cost of self-attention on long sequences by mapping the input to a small latent vector.
ALBERT~\citep{lan2019albert} further uses matrix decomposition to reduce the size of the embedding layer, which is one of the largest consumers of model parameters. 
Finally, \citet{reid-etal-2021-subformer-exploring} studied ways to share weights in transformers, showing that sharing only the middle layers of the model outperforms the alternatives.

\subsection{Retrieval-Augmented Models}

Parametric models can be combined with retrieval mechanisms for text generation, leading to semi-parametric models \citep{gu2018search, lewis2020retrieval,li2022survey}.
This typically amounts to trading model size %
with the number of database entries. For instance, RETRO~\citep{borgeaud2021improving} matched the performance of models 25 times larger, by retrieving chunks of tokens from a 2 trillion token database.
At inference time, the model retrieves tokens / phrases / sentences from a database, which are used by the model through a combination of probability distributions \citep{khandelwal2019generalization}, gating mechanisms \citep{yogatama2021adaptive}, or attention \citep{borgeaud2021improving}.

These models also have good generalization properties: by retrieving from domain-specific databases, they can be applied to new domains, reducing the need for domain-specific fine-tuning \citep{khandelwal2019generalization,khandelwal2020nearest}. 
That is, having an explicit ``memory'' also allows retrieval-augmented models to be adapted post-training. 
Although they may yield slow running speeds as the retrieval time grows as the datastore scales, %
recent works proposed strategies to alleviate this, such as pruning the database~\citep{he2021efficient}, having smaller input-dependent databases~\citep{meng2021fast}, reducing the representation dimension~\citep{martins2022efficient},
and clustering data points \citep{wang2021faster, alon2022neuro}. 
In particular, \citet{martins2022chunk} have shown that carefully constructing a database not only leads to better translations than fine-tuning, but can also reduce the total translation time (inference + online adaptation).\looseness=-1

\subsection{Model Design Considerations}
Despite considerable advances, one major challenge is modeling long sequences in many real-world documents.
For instance, sustainability reports have on average 243.5 pages~\citep{manes2018ensuring} which substantially exceeds the maximum length (16k tokens) found in Path-X from Long Range Arena~\citep{tay2021long}. %
In fact, the ability of a model to handle longer sequences than those seen during training may depend on design choices, such as the attention mechanism~\citep{dubois-etal-2020-location} and the positional encoding~\citep{shaw-etal-2018-self,press2022train}. 
The effect of this behavior when using transformers with sub-quadratic attention, sparse modeling approaches, or parameter efficient models is not yet well-understood.

While sparse modeling approaches like MoE can substantially reduce inference and training costs, they require additional model parameters for retraining specialized modules and have instability issues during training~\citep{zoph2022designing}.
Models that rely on built-in sparse transformations, such as entmax~\citep{peters-etal-2019-sparse}, have achieved strong results without stability issues, but have not yet fully realized competitive efficiency gains. 
Combining MoE with built-in sparse functions may be a promising research direction, e.g., by using entmax in the routing layer.

In retrieval-augmented models, the quality of the retrieval component is critical to performance, and the tradeoff between storing information in model parameters vs.~external resources needs to be better understood, especially when deploying models in low-resource settings like edge devices.
Finally, while new model designs improve efficiency through different means, further improvements can emerge from combining approaches, such as making MoE more efficient using quantization (\cref{sec:quantization}) and using parameter-efficient models for distillation (\cref{sec:distillation}).

\section{Pre-training}\label{sec:pretraining}

Modern transfer learning approaches in NLP %
typically involve \textit{pre-training} a model in a self-supervised fashion on large amounts of text 
before fine-tuning it on specific tasks (\cref{sec:finetuning}). 
Improving the pre-training procedure of a model can significantly reduce the cost of hyperparameter tuning and increase data efficiency for fine-tuning~\citep{Peters:2018,he2019rethinking,neyshabur2020being}.

\subsection{Optimization Objective} 
The choice of the task can determine the success of the pre-trained model on downstream tasks. 
Left-to-right language models, such as GPT~\citep{radford2019language, brown2020language} and PaLM~\citep{palm}, are trained with the \emph{causal language modeling} (CLM) objective, which involves predicting the next token given a context. 
BERT~\citep{Devlin:2019} uses a \emph{masked language model} (MLM) task, which involves filling randomly masked tokens. %

To make better use of available data, various masking strategies have been investigated.
Masking objects and content words only rather than random tokens~\citep{Bitton:2021} or masking more tokens~\citep{wettig2022should} has led to higher task performance and more efficient use of the available data.
ELECTRA~\citep{Clark:2020} and DeBERTa~\citep{he2021debertav3} tried \textit{replaced token detection} (RTD), an objective that uses a small generator model to replace input tokens, and converges more quickly to better performance.
A limitation of the MLM and RTD objectives is that they work with single token replacements.
T5~\citep{raffel2019exploring} and BART~\cite{lewis-etal-2020-bart} overcome this by adopting a denoising \emph{sequence-to-sequence} objective to pretrain an encoder-decoder model, allowing the decoder to predict a span of tokens for masked positions.
In practice, this allows training on shorter sequences without losing task performance, which helps to reduce training costs.

\subsection{Pre-training Considerations}
Despite increases in the size of pre-trained models (cf.~\cref{fig:large_pretrained_models}), many pre-training efficiency gains come from improving model design (\cref{sec:model_design}) and selection (\cref{sec:model_selection}) as well as making more efficient use of the available data (\cref{sec:data}).
These factors have had a greater impact on model performance than the pre-training objective itself~\citep{alajrami-aletras-2022-pre}.
However, pre-training is usually computationally expensive, requiring significant amounts of GPU memory and computational power~\citep{rae2022scaling}, 
and may require large amounts of quality data which can be difficult to acquire and curate~\citep{kaplan2020scaling}. 
Surprisingly, as demonstrated by Chinchilla~\citep{Hoffmann2022}, 
decreasing model size to account for the amount of available data not only leads to better performance, but also reduces computational cost and improves model applicability to downstream tasks. Continued focus on the role of data in efficient pre-training is a promising direction, such as recent work studying the role of (de-)duplication of examples in large scale pretraining corpora \citep{lee-etal-2022-deduplicating}.
While transformers have been the dominant architecture in pre-trained models, more efficient modeling methods such as state space representations and MoEs (\cref{sec:attention}) have the potential to overcome some challenges of pre-training transformers.\looseness=-1

\section{Fine-tuning}\label{sec:finetuning}

\textit{Fine-tuning} refers to adapting a pre-trained model to a new downstream task.
While some approaches explicitly aim to make the fine-tuning process more efficient, 
in this survey, we use a broader definition of fine-tuning that includes any method used to apply a pre-trained model to a downstream task.

\subsection{Parameter-Efficient Fine-Tuning}

Gradient-based fine-tuning typically involves training all model parameters on a downstream task.
Hence, fine-tuning a pre-trained model on a new task creates an entirely new set of model parameters. %
If a model is fine-tuned on many tasks, the storage requirements can become onerous.
Adapting a pre-trained model to downstream tasks by training a new classification layer and leaving the rest of the parameters fixed (aka feature extraction, \citealp{Peters:2018}) updates dramatically fewer parameters than training the full model but has been shown to produce worse performance and has become less common~\citep{Devlin:2019}.

Several approaches have been proposed to adapt a model to a new task while only updating or adding a relatively small number of parameters---up to four orders of magnitude fewer parameters than full-model fine-tuning---without sacrificing (and in some cases improving) performance.
Adapters~\citep{houlsby2019parameter,bapna2019simple,rebuffi2017learning, pfeiffer-etal-2020-adapterhub} inject new trainable dense layers into a pre-trained model, while leaving the original model parameters fixed.
They have recently been improved by the Compacter method~\citep{mahabadi2021compacter}, which constructs the adapter parameter matrices through Kronecker products of low-rank matrices.
While adapters can reduce training time due to a reduced number of trained parameters, and mitigate some deployment costs due to reduced storage requirements, one shortcoming is increased inference time due to more parameters~\citep{ruckle-etal-2021-adapterdrop}.
To mitigate this, \citet{moosavi-etal-2022-adaptable} proposed training an additional layer selector, to only use adapter layers necessary for a given task. %

As an alternative to adding new layers, parameter-efficiency can be achieved by directly modifying activations with learned vectors, either by concatenation~\citep{liu2021gpt, li-liang-2021-prefix, lester2021power}, multiplication~\citep{liu2022few}, or addition~\citep{zaken2021bitfit}.
Two notable approaches are prefix-tuning~\citep{li-liang-2021-prefix} and prompt-tuning~\citep{lester2021power} which fine-tune continuous prompts as an alternative to engineering discrete prompts (cf.~\cref{sec:prompting}).
Although they are conceptually similar to adapters, \citet{he2022towards} show that they are equivalent to a parallel insertion whereas adapters are inserted sequentially.
Alternatively, rather than adding new parameters or changing the computational graph, it is possible to make sparse \cite{sung2021training,guo2021parameter} or low-rank (LoRA, \citealt{hu2022lora}) updates.
Finally, optimization can be performed in a low-dimensional subspace \cite{li2018measuring}, which leads to parameter-efficient updates \cite{aghajanyan2021intrinsic}.
Although low-rank approaches mitigate the issue of increased inference time, they require an additional optimization step to identify the best rank. 
To mitigate this, \citet{valipourdylora} proposed a dynamic solution that substantially reduces training time compared to LoRA.
Lastly, \citet{wang-etal-2022-adamix} devised AdaMix to combine different parameter efficient fine-tuning techniques together via routing and showed that their approach can even outperform full fine-tuning.

\subsection{Multi-Task and Zero-Shot Learning}

While traditional transfer learning includes fine-tuning, there are other paradigms that allow for immediate application of a pre-trained model to a downstream task of interest.
\textit{Multi-task learning} \cite{caruana1997multitask,ruder2017overview} aims to train a single model that can perform a wide variety of tasks out of the box. Typically, this is done by fine-tuning on data from all downstream tasks of interest.
Multi-task models can improve fine-tuning performance \cite{raffel2019exploring,aghajanyan2021muppet,aribandi2021ext5,liu2022few}.
In certain cases, a multi-task model works on new tasks without any fine-tuning, also referred to as \textit{zero-shot generalization} \cite{sanh2022multitask,wei2022finetuned}.
\citet{radford2017learning,radford2019language} and \citet{brown2020language} demonstrated that language models trained with an unsupervised objective can perform a variety of tasks out-of-the-box.
While it can circumvent the need for fine-tuning, zero-shot ability depends on model size and only becomes competitive at a certain scale \citep{wei_emergent_2022}.

\subsection{Prompting}\label{sec:prompting}
Inspired by models like GPT-3~\citep{brown2020language}, prompting refers to casting a task as a textual instruction to a language model~\citep{liu2021pre}. 
In general, prompts can be either crafted manually or automatically using fill-in templates for token, span, and sentence-level completion~\citep{petroni-etal-2019-language,brown2020language,shin-etal-2020-autoprompt}. 
This makes prompting applicable to more challenging NLP tasks, such as QA, MT, and summarization~\citep{schick-schutze-2021-just}.
Although prompting eliminates the need for any fine-tuning, identifying good prompts can be difficult~\cite{liu2021gpt}.
Hence, recent works investigate the automated creation of suitable prompts, albeit with additional training cost~\citep{Bach2022-di}\looseness=-1.

\subsection{Fine-Tuning Considerations}
An emerging problem with large language models is the universally high cost of fully fine-tuning them~\citep{chen2021evaluating}.
Although prompting (without fine-tuning) can alleviate this issue, designing prompts can be tedious---even with automated help.  
One promising direction for efficiently introducing new knowledge into models is to combine existing methods for efficient fine-tuning.
This could involve methods such as~\citet{karimi-mahabadi-etal-2022-prompt}, who proposed task-specific adapters to avoid generating prompts, and achieved considerable speed ups while tuning under 1\% of parameters.
Another challenge in adopting large pre-trained models for fine-tuning is the complexity in interpreting the final model, due in part to the use transformers.
To gain a better understanding of these models while still leveraging efficiency, a premising direction is to combine techniques such as sparse modeling and parameter-efficient methods~\citep{correia_adaptively_2019,treviso-etal-2022-predicting}.

\section{Inference and Compression} \label{sec:inference}

\textit{Inference} involves computing a trained model's prediction for a given input.
Inference can be made more efficient by accelerating the process for time efficiency (latency), or by compressing the model to reduce memory requirements.

\subsection{Pruning}\label{sec:pruning}
Proposed by \citet{LeCun:1989}, pruning removes irrelevant weights from a neural network to reduce computation, and furthermore, decreases memory capacity and bandwidth requirements. 
Pruning can be applied at different stages of the NLP pipeline (\cref{fig:overview}).
For instance, \citet{gordon_compressing_2020} found that up to $\sim$$40\%$ of BERT can be pruned at pre-training without affecting its performance. 
Others proposed pruning methods that work as regularizers and can be applied to pre-training and fine-tuning~\citep{louizos2018learning, wang-etal-2020-structured}.
Finally, works investigated pruning during fine-tuning~\citep{han2015learning,sanh_movement_2020} or dynamically during inference~\citep{Fan:2020}.

Pruning was initially introduced at the individual weight level (unstructured pruning), but more recent approaches prune larger components of the network (structured pruning).
Examples of the latter include removing attention heads~\citep{voita_analyzing_2019,michel_are_2019},  weak attention values~\citep{ji_distribution_2021,qu_dota_2022}, and even entire hidden layers~\citep{dong2017learning, sajjad_poor_2020}. 
In particular, \citet{xia-etal-2022-structured} found that pruning all these components yields more accurate and efficient models.
When comparing the two pruning approaches, unstructured pruning is often found to better preserve a model's performance \citep{gale2019,ahia-etal-2021-low-resource}, but existing hardware often cannot exploit the resulting sparsity. In contrast, structured pruning methods often lead to a higher improvement in terms of inference speed~\citep{hoefler2021sparsity}.
The increasing popularity of pruning methods has further raised the question of how to quantify and compare them~\citep{gale2019,blalock2020state,kaleab2021,hoefler2021sparsity} and motivated works that combine pruning with other efficiency methods such as adapters~\citep{ruckle-etal-2021-adapterdrop} and distillation~\citep{zafrir2021prune}.

While early pruning (e.g., during pre-training) can further reduce training costs, it increases the risk of over-pruning: removing nodes essential for downstream task performance~\citep{gordon_compressing_2020}.
Although this can be mitigated by ``regrowing" pruned weights~\citep{ICML-2019-MostafaW}, this increases training costs.
Other pruning downsides include additional costs for hyperparameter tuning such as the number of preserved weights.

\subsection{Knowledge Distillation}\label{sec:distillation}
The process of knowledge distillation uses supervision signals from a large (teacher) model to train a smaller (student) model~\citep{hinton-2015}, and often leads to the student outperforming a similarly-sized model trained without this supervision.
While early works focused on distilling task-specific models~\citep{kim-rush-2016-sequence}, recent works focus on distilling pre-trained models that can then be fine-tuned on specific downstream tasks~\citep{sanh_distilbert_2019,liu-etal-2020-fastbert,jiao-etal-2020-tinybert,sun-etal-2020-mobilebert,Gou:2021}.
The downsides of distillation include the added cost of tuning student hyperparameters and the potential for reduced performance and generalization capability~\citep{stanton2021does}. Recently, \citet{zhu2022teach} discovered that some performance loss is due to undistillable classes and suggested ways to address this.

\subsection{Quantization}\label{sec:quantization}
Mapping high-precision data types to low-precision ones is referred to as \textit{quantization}.
Quantization can be applied at different stages in the NLP model-building pipeline to reduce training and inference costs.
Various research has shown that low-precision data format can reduce memory consumption by 4x--24x and improve the throughput by 4.5x compared to 32-bit floating point format.
Various works targeted specific precision-levels such as integers~\cite{kim_i-bert_2021}, 8-bit~\cite{quinn-ballesteros-2018-pieces, zafrir_q8bert_2019, bhandare_efficient_2019, prato_fully_2020, dettmers2022llm}, ternary~\cite{zhang_ternarybert_2020, ji_distribution_2021,zadeh_mokey_2022}, and even binary representations~\cite{bai_binarybert_2020}.

Different components may have a different sensitivities regarding their underlying precision, so there is a body of work on mixed-precision quantization. 
\citet{shen_q-bert_2020} showed that embedding layers require more precise parameter representations than the attention layer, while \citet{kim_i-bert_2021} showed that nonlinear functions require more bits than the general matrix multiplication.
Others defined quantization as a constrained optimization problem to automatically identify layers where lower precision is sufficient~\citep{pmlr-v139-hubara21a}.
Finally, several works proposed quantization during training to make them robust against performance loss after quantization~\citep{zafrir_q8bert_2019,kim_i-bert_2021,stock2021training}.
For instance,~\citet{bai_binarybert_2020} and \citet{zhang_ternarybert_2020} proposed using knowledge distillation to maintain the accuracy of binarized and ternarized models.
These show that component-customized quantization can preserve accuracy while improving efficiency.
To maximize the benefit from quantization, one should also consider the available underlying hardware and associated specialized kernels compatible with different bit representations~\citep{noune20228, kuzmin2022fp}.

\subsection{Inference Considerations} 

While efficiency during pre-training and fine-tuning focuses on the computational resources and time required to train and optimize a model, inference efficiency is focused on how well a learned model can perform on new input data in real-world scenarios.
Moreover, inference optimization is ultimately context-specific and the requirements vary according to the use-case. Therefore, there is no one-size-fits-all solution to optimizing inference, but instead a plethora of techniques.
For instance, while \citet{wu2022extreme} combine several methods to achieve utmost model compression, other works improve task-specific mechanisms such as beam-search in MT~\citep{peters-martins-2021-smoothing}. 
Parallelism can also be leveraged to increase inference efficiency, but its effectiveness may depend on the hardware available~\citep{Rajbhandari2022}.
Dynamic computation techniques, such as early-exit~\citep{schwartz-etal-2020-right,xin-etal-2020-deebert} and mixture-of-experts~(\cref{sec:attention}), can improve inference efficiency by selectively performing computation only on the parts of the model that are needed for a given input. 
However, current dynamic computation methods often use eager execution mode, which can prevent them from low-level optimization, as noted by~\citet{xu2022survey}. Work focusing on inference efficiency should carefully report the exact target setting (hardware, eager vs.~static execution framework). Accordingly, promising directions for optimizing inference efficiency might consider tighter integration across or more general purpose approaches with respect to algorithm, software and hardware. One recent such example is neural architecture search for hardware-specific efficient transformers~\citep{wang-etal-2020-hat}.

\section{Hardware Utilization}\label{sec:hardware}

Many hardware-specific methods focus on reducing GPU memory consumption, a major bottleneck in transformer models. 
Others leverage specialized hardware, co-design of hardware, and adaptations targeted to edge devices.
Many techniques can be combined and applied across different stages of training and inference (\cref{fig:overview}) for further efficiency.

\subsection{Reducing Optimizer Memory} 

Optimizers that track gradient history incur a memory cost.
Libraries like DeepSpeed~\cite{ren_zero-offload_2021} allow gradient history to be offloaded from GPU to CPU RAM where computation is performed via efficient AVX instructions. 
\texttt{bitsandbytes}~\citep{dettmers2022bit} uses dynamic block-wise quantization to reduce memory pressure. It splits tensors into blocks and quantizes each block individually. This reduces memory consumption by 75\% and improves training times due to reduced inter-GPU communication.

\subsection{Specialized Hardware} \label{sec:spec-hardware}
Specialized NLP hardware has been built using Application Specific Integrated Circuits or Field Programmable Gate Arrays, though it is not yet broadly available.
These designs use dedicated units for efficient operations like quantization and pruning (\cref{sec:inference}). %
For example, \citet{zadeh_gobo_2020, zadeh_mokey_2022}, \citet{li_efficient_2021}, and \citet{qu_dota_2022} support ultra-low-bit and mixed precision computation that cannot be done on CPUs/GPUs;
\citet{ham_3_2020,ham_elsa_2021} and \citet{wang_spatten_2021} design hardware that predicts and prunes redundant heads/tokens and weak attention values in transformers. \citet{qu_dota_2022} presents a design that balances the workload to alleviate the irregularity in the pruned attention.
Others develop new types of processors and memories optimized for transformer components: 
\citet{lu_hardware_2020} and \citet{liu_hardware_2021} implemented dedicated hardware for softmax and layer normalization respectively, and \citet{tambe_edgebert_2021} used embedded Resistive RAM---a nonvolatile memory with low latency and energy consumption---to store word embeddings.\looseness=-1

\subsection{Co-design}
Some works optimize hardware, software, and algorithms jointly, which historically has been a common way to realize efficiency gains~\citep{Hooker2021}.
For instance,~\citet{Lepikhin2021GShardSG} demonstrated that improving the underlying compiler can substantially improve parallelization and enable scaling.
Other examples for co-design focus on hardware-aware mixture of experts models and attention mechanisms to produce substantial speedups~\citep{Jiaao2022,Rajbhandari2022,Dao2022-yl}. %
\citet{Barham2022} proposed a gang-scheduling approach with parallel asynchronous dispatch, that leads to substantial efficiency gains.
Finally,~\citet{hinton2022forward} suggested ``mortal computation'', an extreme form of co-design, where by training a model that is tailored to a specific hardware, the need to guarantee consistent software behavior across different hardware is reduced, potentially saving computation.
 
\subsection{Edge Devices}
Tight compute and memory constraints on edge devices motivate a separate set of efficiency solutions.
SqueezeBERT~\cite{iandolaSqueezeBERTWhatCan2020a} incorporates group convolutions into self-attention to improve efficiency on mobile devices. 
EdgeFormer~\cite{geEdgeFormerParameterEfficientTransformer2022} interleaves self-attention layers with lightweight feed-forward layers and an encoder heavy parameterization to meet edge memory budgets.
GhostBERT~\cite{huangGhostBERTGenerateMore2021} uses \textit{ghost} modules built on depth-wise separable convolutions used in MobileNets~\cite{howardMobileNetsEfficientConvolutional2017}.
LiteTransformer~\cite{wu*LiteTransformerLongShort2022} uses long-short range attention to encode local context by convolutions for MT in resource-constrained settings. %
Through quantization \texttt{llama.cpp}\footnote{https://github.com/ggerganov/llama.cpp, 20 March 2023} runs a 7B-parameter LLM on recent mobile phone hardware.
Finally, ProFormer~\cite{sankarProFormerOnDeviceLSH2021} reduces runtime and memory via locality sensitive hashing and local projection attention layers.\looseness=-1

\subsection{Hardware Considerations}

To deliver more computational power, vendors pack denser computational units into domain-specific hardware, such as tensor cores in  Intel FPGAs, Xilinx AI Engines, and  matrix processors in the Google TPU.
However, irregularities in the transformer, like sparsity and mixed data types, restrict the use of these  resources.
We suggest focusing on adapting efficient transformers to existing specialized hardware platforms, including using hardware-optimized data formats like block floating point, and exploring sparsity on dense tensor units.

\section{Evaluating Efficiency} \label{sec:evaluation}

Evaluating efficiency requires establishing which computational aspect one aims to minimize.
We discuss the two most prominent aspects (FLOP/s and power consumption), and list open challenges.

\subsection{Evaluation Measures} \label{subsec:measuring}
\paragraph{Pareto optimality}
When improving efficiency, multiple factors often need to be traded-off. 
For instance, longer training time can increase task performance, but simultaneously increase resource consumption.
A principled way to characterize trade-offs is to identify Pareto-optimal solutions~\citep{pareto1896cours}, %
those for which no other system reaches a better or equal task performance with lower resource consumption. 
As there may be several Pareto-optimal solutions, final choice depends on the application context; a small, average-quality model and a large, higher-quality model can both be optimal. 
Thus, as long as a model contributes to or extends the Pareto-optimal curve for a given problem and measurement space, it it worthwhile---even if other solutions may use less resources or produce higher quality scores.

Advancing NLP by pushing Pareto barriers is an established practice~\citep{kim-etal-2019-research,bogoychev-etal-2020-edinburghs,behnke-heafield-2021-pruning}. 
For instance, the WNGT 2020 MT shared task~\citep{ngt-2020-neural} considers the Pareto frontier between real time taken, system or GPU memory usage, and model size, as well as BLEU score. %
\citet{de2021hyperparameter} included power consumption as a trade-off against perplexity to explore Pareto-efficient hyperparameter combinations for transformer models. 
Finally, \citet{liutowards} examined Pareto efficiency for a number of tasks in an attempt to narrow model selection search space.

\paragraph{FLOP/s} 
A frequently reported efficiency measure is the number of floating point operations (FLOPs) and floating points per second (FLOP/s).
While these discrete metrics seem well-defined in terms of what the hardware does, there is some variation at multiple stages of the stack, adding uncertainty. 
For example, different operations may count as a FLOP on different hardware; non-floating-point operations are not considered; and hardware is rarely 100\% utilised and achieving this productively is a challenge, so theoretical FLOP/s performance cannot be multiplied with time elapsed to yield the amount of computing performed. 
Still, FLOP/s per unit power can indicate which hardware choices have the potential to offer Pareto-efficient trade-offs~\cite{hsu2005towards}.

\paragraph{Power consumption} 

There exist various ways to measure power consumption, for instance, by using specific hardware such as an electricity meter.
While this can provide precise figures with a high temporal accuracy, it cannot provide a fine-grained estimate for individual computers in a network.
Moreover, it does not cover external energy costs such as cooling or networking.
Another way is to use software tools such as MLCO2~\citep{lacoste2019quantifying}. 
Some tools even provide a real-time breakdown of the power consumption of different components within a machine~\citep{henderson2020towards} or local machine API-reported figures to stop training early if prudent~\citep{anthony2020carbontracker}.
Finally, \citet{hershcovich2022towards} introduced a model card for NLP systems that encourages researchers to document efficiency in a consistent manner.

Measuring power consumption programmatically comes with a number of caveats. 
First, sampling frequency is often restricted at various levels of the stack and may result in a lag in measurement start. 
Consequently, shorter experiments may log an energy use of zero, and there will almost always be energy demand that is missed. 
Second, inefficiencies such as heat loss are not reported by current APIs and hence, do not cover cooling and other system management activities. 
Third, not all architectures and operating systems are supported.
For instance, power consumption under macOS is difficult to manage, and direct figures for TPU power consumption are not available.

\paragraph{Carbon emissions}
Carbon emissions %
are usually computed using the power consumption and the carbon intensity of the marginal energy generation used to run the program. 
Thus, low-energy does not mean low-carbon, and high-energy models can---in the right region and with some care---be zero-carbon in terms of point energy consumption impact, if executed at the right time (i.e., when the energy mix is low-carbon intensity, \citealp{Dodge:2022}). 
For estimating the CO$_2$ emissions from a specific program execution, APIs such as ElectricityMap\footnote{\url{https://electricitymap.org}} provide real-time access to carbon intensity for many regions. 
However, as carbon intensity varies and is affected by other factors like the power usage efficiency in a data center, it is often a poor basis for comparison; in fact, \citet{henderson2020towards} recommended using multiple runs for a stable estimate.
Furthermore, one needs to consider that zero-carbon program executions still consume energy, and that efficiency does not intrinsically guarantee a reduction in overall resource consumption, as the resulting cost reduction may lead to an increase in demand counteracting any gains, an effect known as Jevons' paradox~\cite{jevons1866coal}.

\subsection{Open Challenges in Measuring Efficiency}

Hardware choice can lead to pronounced differences in certain efficiency measurements such as latency and thoroughput~\citep{LeeThorp2021FNetMT}. 
Properly measuring efficiency remains a major challenge \citep{cao-etal-2020-towards}. 

\paragraph{Separating different stages} 
It is important to characterize efficiency of pre-training and fine-tuning stages separately (\cref{sec:pretraining,sec:finetuning}).
Models may present different memory requirements during training yet result in trained models with comparable inference memory consumption. 
This is because training often involves design choices that increase the memory overhead of backward propagation. 
Further, some optimizers may require substantially more memory than others. 
Similarly, parameter sharing techniques may show little benefits during training but show memory improvements at inference~\citep{dehghani2021}. Finally, while larger models run slower than smaller ones, they converge faster and better compress using methods like pruning and quantization~\cite{pmlr-v119-li20m}.

\paragraph{Disagreement between cost factors}
As partially discussed in \cref{sec:spec-hardware}, cost indicators may disagree with each other.
For instance, mixture of experts increases the overall parameter count, but improves the trade-off between quality and FLOPs, as they minimize the per-data cost by routing to subsections of the model~\citep{Rajbhandari2022}. 
Conversely, unstructured sparsity techniques can significantly minimize the overall number of FLOPs, yet in practice, it introduces low-level operations that can lead to far higher memory requirements to store the indices that indicate what part of the matrix is sparse~\citep{qu_dota_2022}. 
Finally, \citet{dao2021} and \citet{dao2022monarch} found specific sparsity patterns that achieve more predictable speedups with current hardware.\looseness=-1

\paragraph{Trade-offs with other desiderata}
One major, but seldom studied concern when improving efficiency are trade-offs with other desiderata such as fairness and robustness.
For instance, \citet{Hooker2020}, \citet{renduchintala-etal-2021-gender}, and \citet{silva-etal-2021-towards} found that compression techniques such as pruning can amplify existing biases; \citet{Mohammadshahi2022} and \citet{ogueji-etal-2022-intriguing} further explored these trade-offs in a multilingual setting.
So far, only a few works investigated preserving a model's fairness when increasing its efficiency.
To quantify such effects, \citet{xu-etal-2021-beyond} proposed a novel metric called loyalty, which measures the resemblance of predicted distributions made by teacher and student models. \citet{hessenthaler-etal-2022-bridging} established that many approaches for increasing fairness in NLP models also increase computation, and jointly with work like \citet{wang2022} showed that distillation can decrease model fairness.
\citet{xu2022can} studied these effects more systematically, with mixed conclusions. 
While more positive insights have been found with respect to other desiderata such as out-of-distribution (OOD) generalization~\citep{ahia-etal-2021-low-resource,iofinova2021well,ogueji-etal-2022-intriguing} and model transfer~\citep{gordon_compressing_2020}, more work is needed to better understand and benchmark the impact of efficiency beyond accuracy.

\section{Model Selection} \label{sec:model_selection}
Finally, we discuss lines of research that opt to efficiently select a well-performing model variant.

\subsection{Hyperparameter Search} 
The performance of machine learning methods can be improved by choosing hyperparameters carefully. 
Model-based techniques such as Bayesian optimization (BO; \citealp{snoek2012practical,feurer2015efficient}) and graph-based semi-supervised learning \citep{zhang-duh-2020-reproducible} use surrogate models to search efficiently for optimal hyperparameters, avoiding inefficient grid search or manual tuning.
A complementary approach is 
successive halving  (SHA;  \citealp{jamieson2016non}) and its massively parallel variant, asynchronous SHA (ASHA; \citealp{li2020system}), which test multiple hyperparameter settings in parallel for a fixed number of training iterations, then discard the half of the settings with the worst validation set performance. 

The SMAC3 library \cite{lindauer2022smac3} implements several BO strategies, including a budget-limited variant for expensive deep learning tasks, and is integrated into \emph{auto-sklearn} \citep{feurer2020auto} and \emph{auto-pytorch} \citep{zimmer2021auto}.
However, with limited computational budgets, both BO and ASHA may fail to identify good settings \cite{liu-wang-2021-empirical}. 
It is unclear whether these methods can be used to choose random initial weights or to order training samples, which also affect model performance \cite{Dodge:2020}.

\subsection{Hyperparameter Transfer} 
To minimize the number of trials needed to find optimal hyperparameter settings, one can transfer knowledge from other datasets or tasks---similar to how an ML engineer might select reasonable settings by hand. 
Transferring hyperparameters can be especially beneficial during expensive stages in the NLP pipeline, such as pre-training.
Transfer neural processes \cite{wei2021meta} provide a way to transfer observations, parameters and configurations from previous tasks 
using Bayesian optimization with a neural process as the surrogate model. 
This can lead to more accurate models with fewer trials than conventional BO approaches, but has yet to be tested for large NLP models.
Finally, the cost of training can be reduced using $\mu$Transfer \cite{yang2021tuning}, which tunes a small model, then transfers the hyperparameters to a larger model.

\subsection{Model Selection Considerations}
While identifying an optimal model is  crucial in deployment, it  raises several challenges around reporting practices~\citep{reimers-gurevych-2017-reporting,NEURIPS2021_f514cec8} and hyperparameter tuning~\citep{bouthillier2020survey, gundersen2022sources}.\footnote{E.g., when considering  compute budget variation when comparing new model development to baselines.}
A first step towards improved comparability could be to fix hyperparameter tuning  budget~\citep{Dodge:2019,Hoffmann2022},
or consider the full search space \cite{bell2022modeling}.
\section{Conclusion}
This survey provides a broad overview of considerations for increasing efficiency in modern NLP models, identifying both immediate successes and remaining challenges. %
Most progress so far has been in model design, typically targeted at a specific computational budget and hardware paradigm.
Key challenges include better understanding and modelling 
trade-offs between end-task performance and resource consumption, %
and the dependency between hardware choices and software implementations. 
Furthermore, we note that efficiency in NLP has many definitions and can be achieved in many different ways, but is also subject to various open challenges, and cannot be measured by a single metric.
We outline several promising research directions aligned with overcoming these challenges, ranging from approaches that make better use of available data, strategies for reducing the cost of pre-training and fine-tuning large models, to prioritizing the importance of interactions between algorithms, software,  and hardware.

Impressive advances in NLP enabled primarily by scaling computation have produced remarkable progress in a short span of time. However, in order to realize the full potential of this technology for a broader swath of society, we must reduce the amount of computation that is required to achieve these remarkable results. We hope that this survey can serve to accelerate advances in this important area of research with great potential for impact both within our field and for society as a whole.

\section*{Acknowledgements}
This work was initiated at and benefited substantially from the Dagstuhl Seminar 22232: \textit{Efficient and Equitable Natural Language Processing in the Age of Deep Learning}.
We further thank Yuki Arase, Jonathan Frankle, Alexander Koller, Alexander L{\"o}ser, Alexandra Sasha Luccioni, Haritz Puerto, Nils Reimers, Leonardo Riberio, Anna Rogers, Andreas R{\"u}ckl{\'e}, Noah A. Smith, and Thomas Wolf for a fruitful discussion and helpful feedback at the seminar. 
M.T. and A.M acknowledge the European Research Council (ERC StG DeepSPIN 758969), EU’s Horizon Europe Research and Innovation Actions (UTTER, contract 101070631), and Fundação para a Ciência e Tecnologia through contract UIDB/50008/2020.
L.D. acknowledges support of the Independent Research Fund Denmark under project 9131-00131B, Verif-AI, and the Novo Nordisk Foundation project ClinRead, NNF19OC0059138.
Finally, we also thank the TACL reviewers and action editor for helpful discussion and insightful feedback.

\bibliography{custom, hardware_accelerators,edge}
\bibliographystyle{acl_natbib}

\end{document}